\documentclass[sigconf]{acmart}

\usepackage{amsmath,amssymb,amsfonts}
\usepackage{algorithmic}
\usepackage{tabularx}
\usepackage{graphicx}
\usepackage{textcomp}
\usepackage{multirow}
\newcolumntype{Y}{>{\centering\arraybackslash}X}

\AtBeginDocument{%
  \providecommand\BibTeX{{%
    \normalfont B\kern-0.5em{\scshape i\kern-0.25em b}\kern-0.8em\TeX}}}

\setcopyright{acmcopyright}
\copyrightyear{2019}
\acmYear{2019}

\acmConference[DTMBio \textquotesingle19]{DTMBio 2019: ACM 13th International Workshop on Data and Text Mining in Biomedical Informatics}{November, 2019}{Beijing, China}

\begin{document}

\title{Deep User Identification Model with Multiple Biometrics}

\author{Hyoung-Kyu Song}
\affiliation{%
  \department{Research Institute}
  \institution{NOTA Incorporated}
}
\authornote{These are contributed equally in this research.}

\author{Ebrahim AlAlkeem}
\affiliation{%
  \department{EECS Dept.}
  \institution{Khalifa University}
  }
\authornotemark[1]

\author{Jaewoong Yun}
\affiliation{%
  \department{Research Institute}
  \institution{NOTA Incorporated}
}

\author{Tae-Ho Kim}
\affiliation{%
  \department{Institute for Artificial Intelligence}
  \institution{Korea Advanced Institute of Science and Technology}
}
\author{Hyerin Yoo}
\affiliation{%
  \department{Research Institute}
  \institution{NOTA Incorporated}
}
\author{Dasom Heo}
\affiliation{%
  \department{Research Institute}
  \institution{NOTA Incorporated}
}

\author{Chan Yeob Yeun}
\affiliation{%
  \department{C2PS, EECS Dept.}
  \institution{Khalifa University}
}

\author{Myungsu Chae}
\affiliation{%
  \department{Research Institute}
  \institution{NOTA Incorporated}
}
\authornote{mschae89@nota.ai (corresponding author)}

\renewcommand{\shortauthors}{H.-K. Song and E. Al Alkeem, et al.}

\begin{abstract}

Identification using biometrics is an important yet challenging task. Abundant research has been conducted on identifying personal identity or gender using given signals. Various types of biometrics such as electrocardiogram (ECG), electroencephalogram (EEG), face, fingerprint, and voice have been used for these tasks. Most research has only focused on single modality or a single task, while the combination of input modality or tasks is yet to be investigated. In this paper, we propose deep identification and gender classification using multimodal biometrics. Our model uses ECG, fingerprint, and facial data. It then performs two tasks: gender identification and classification. By engaging multi-modality, a single model can handle various input domains without training each modality independently, and the correlation between domains can increase its generalization performance on the tasks.

\end{abstract}




\maketitle

\section{Introduction}
Recognition is an essential function of human beings. Humans easily recognize a person using various inputs such as voice, face, or gesture. Thus, when engaging with deep learning, multi-modality needs to be taken into account instead of single modality. 

Using multi-modality has many benefits including noise reduction. As any single modality can be easily contaminated due to powerline, EMG, channel noise, electrodes, and motion artifact, it is difficult to build a single modal recognition algorithm. For multimodal model, if a modality has a low signal to noise ratio, it can be compensated by another modality, so the overall system performance is maintained. Additionally, in the training phase, correlated features among modalities can be trained, and this may lead to a better performance. Moreover, when compared to training three independent models on each modality, it is more efficient to train a single multimodal model.

In this study, we mainly focus on DL model to get better result. Previously, traditional machine learning methods were engaged for person identification. However, ML contains pipeline of hand-crafted feature extraction followed by classification model such as k-NN, random forest, and etc. Since the feature is hand-crafted, there is no guarantee that only informative features are extracted. Using DL, the choice of feature extraction and the learning model is far less restricted comparing to ML.

In the real world, the ECG from a person may vary because of several factors such as heart rate, sickness, etc. If the model is trained using multitask learning, the model can focus only on the target task, and ignore factors that are of no interest. In addition, there may be noise in the procedure to collect data. Therefore, it is necessary to develop a robust model architecture, considering generalization.

In this paper, we introduce our user identification model which uses three biometrics: ECG, face, and finger. Moreover, this model is designed for solving supplementary tasks such as gender classification. Finally, to use multimodal biometrics, it fuses features from each biometric and is trained with a deep neural network; thus, it can be robust on noisy data.

\begin{figure*}[t]
  \centering
  \includegraphics[width=0.7\linewidth]{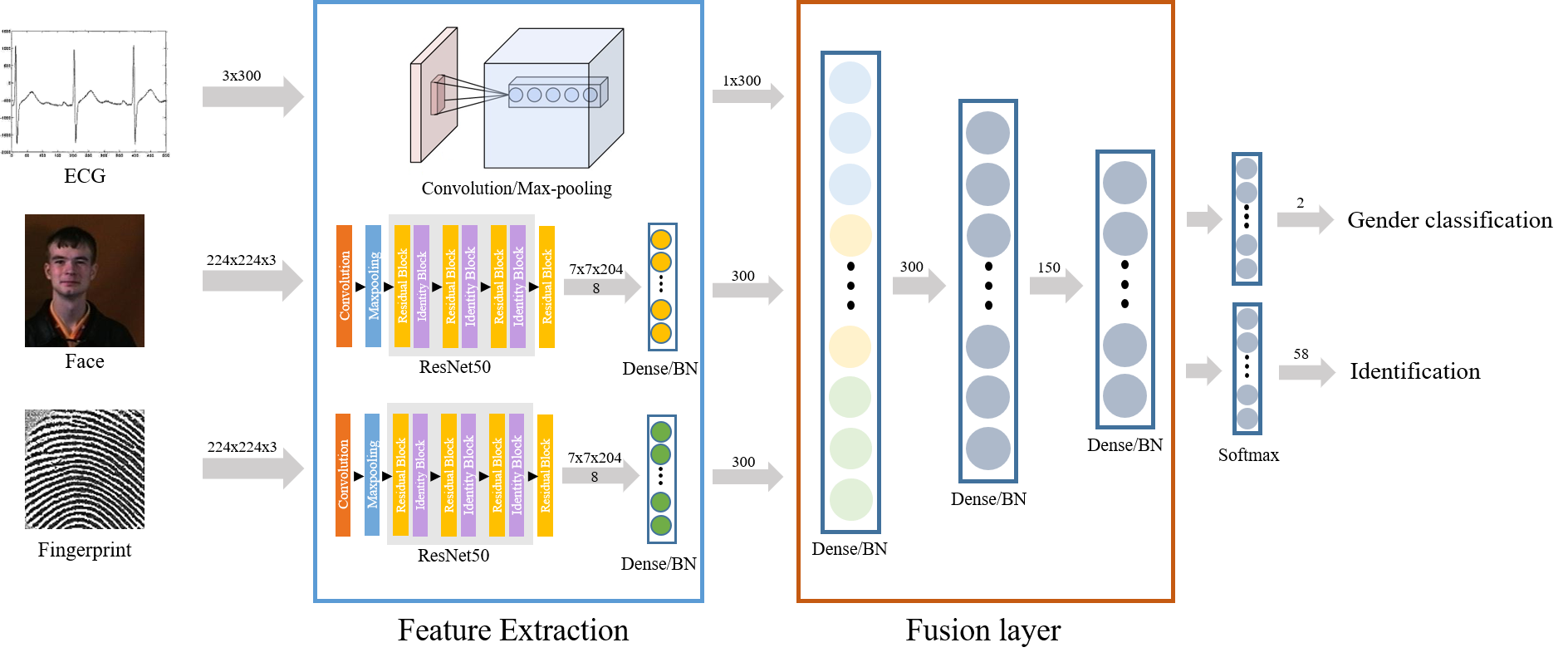}
  \vspace{-0.4cm}
  \caption{The proposed network architecture}
  \vspace{-0.3cm}
  \label{fig:model}
\end{figure*}

\section{Related Works}

Recent studies have proposed new approaches to enhance system security. They have adopted the method of combining two or more biometrics to identify or certify users. These techniques have demonstrated that extracting features from multi-modality and fusing them is effective in increasing the accuracy of the verification \cite{b1}. Sara and Karim emphasized the effects of adopting multi-modality in their work, which showed that the performance was raised from 82.1\% (only palmprint) and 89\% (only ECG) to 94.7\% (multi-modality; palmprint and ECG) \cite{b2}.

A virtual multimodal database is widely used in multimodal biometrics research \cite{b3, b4, b5}. A virtual dataset can reduce wastage of time and the cost of data collection. Let us assume that we have two databases which are mutually exclusive on modality and subjects. Then, the virtual multimodal database is constructed by matching subjects in the two databases to allow a single person to obtain information on both modalities. This could be performed based on the assumption that different biometric traits of the same subject are independent \cite{b6}. There is no public multimodal database which consists of ECG, face, and fingerprints extracted from the same person.
\vspace{-0.2cm}

\section{Methods}

As shown in Fig. \ref{fig:model}, the network architecture is largely divided into three parts: feature extraction, which transforms the input into an embedding space; fusion layer, which combines features from each modality; and task layer, which performs the task. First, feature extraction is performed on each modality using a different method as each modality has different characteristics. The datatype of fingerprint $x_p$ and face $x_f$ is a static image while ECG $x_e$ is a temporal biological signal. It is well known that a classifier trained on large database can be used as a general feature extractor \cite{b7}.

We used ResNet-50 for extracting features from $x_f$ and $x_p$. Additionally, a fully connected layer is used for matching the number of dimensions to that of ECG. While CNN architecture followed by max pooling is designed for $x_e$ for ECG, the extracted feature does not rely on the temporal axis. Second, the fusion layer takes concatenated features from the feature extraction module as an input. We use neural networks for the fusion layer. As the data propagates, the information of each modality is mixed up to decrease a loss. Finally, the task layer takes the output of fusion layer as input.

The model calculates the loss with categorical cross-entropy for user identification, and binary cross-entropy for gender classification. For multitask experiment, the joint loss is decided from these two losses by adding with same weight.

\subsection{ECG Preprocessing}

We cropped a total of 300 data values before and after R peak of the signal to generate a QRS complex vector. After this procedure, min-max normalization is applied to each QRS complex. We can get at least 15 QRS complexes from each record. As mentioned earlier, because we have the intuition that temporal information between QRS complexes is less important, we extract three QRS complexes for one input from a record. In conclusion, one input sequence has three timesteps, which means that three QRS complexes are grouped to an input sequence.

\subsection{Feature Extraction}

Residual Network (ResNet \cite{b8}) is a widely used feature extractor for images. ResNet is trained on the ImageNet database, which is the largest database for object classification. We utilize ResNet50 as a feature extractor for face and fingerprint. We use average pooling of feature (7 x 7 x 2048).

Although there is a model based on the LSTM architecture for user identification \cite{b9}, we have used the CNN architecture as a feature extractor with the intuition that temporal information between QRS complexes is less important. The CNN model has a 1D convolution layer which takes a (batch size, time step, 300) tensor as input and outputs a (batch size, time step, 300) tensor. After the 1D convolution layer, 1 max pooling operation is used. The extracted ECG biometric feature is concatenated with 2 feature vectors from other modalities.

\subsection{Feature Fusion}

There are many fusion methods. In terms of improving the model performance, it is very difficult to properly fuse three different modalities at the input level. We compare the performance between score level fusion and feature level fusion. The three methods of score level fusion, which are sum, product, and max rule, were tested. At the feature level, the three extracted features from each modality are respectively normalized and then concatenated as the input for the model. In our model design, when the specific biometric is not given, the feature value for corresponding modality is processed as 0 and excluded from BN layers. Thus, we can change the number of modalities used in a single model.

\begin{figure}[t]
  \centering
  \includegraphics[width=0.97\linewidth]{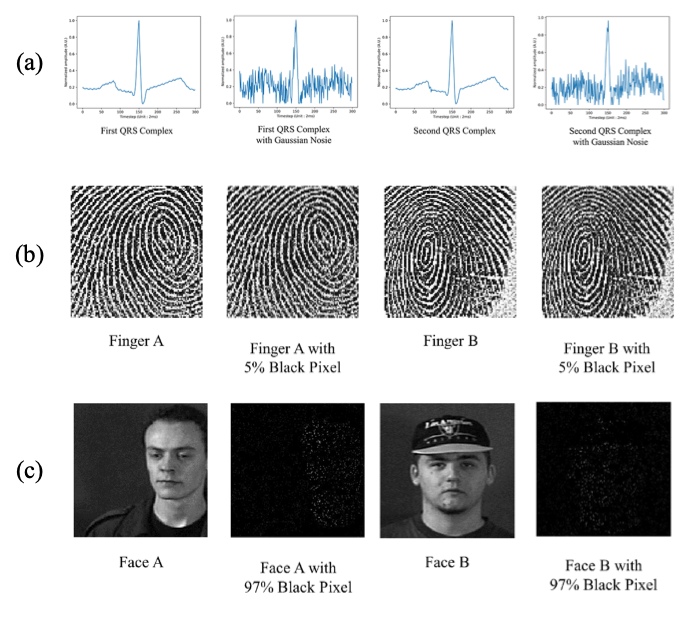}
  \vspace{-0.6cm}
  \caption{Raw and noisy sample of (a) ECG, (b) fingerprint image, and (c) facial image.}
  \vspace{-0.4cm}
  \label{fig:noisy}
\end{figure}

\section{Experiments}

This section presents our experiments for verifying the superiority of the proposed model using the accuracy metric. Three experiments were conducted in this research. The first experiment compared single modality and multimodality for multitask learning (gender classification and user identification). The second experiment was a comparison between multitask learning and single task learning by the multimodal biometrics, which consisted of ECG data (ECG-ID database \cite{b10, b11}, PTB database \cite{b10, b12}), face data (Face95 \cite{b13}), and fingerprint data (FVC2006 \cite{b14}). Finally, we add noise to the data to verify its robustness. Especially, we assumed that the face and fingerprint are used as supplementary biometrics, so we introduced severe noise to these two modalities.

\subsection{Dataset}
\subsubsection{ECG} 
For ECG, we combined two different datasets, ECG-ID database and PTB database, both of which can be downloaded from Physiobank in Physionet. By using multiple datasets for the biometric, the model can work regardless of where the signals are measured. The ECG-ID database contains 310 recordings from 90 subjects (44 males and 46 females, aged from 13 to 75 years). Each recording was measured by Lead I, recorded for 20 seconds, and digitized at 500 Hz with 12-bit resolution over a nominal 10 mV range. The PTB ECG database contains 549 records obtained from 290 subjects (209 males and 81 females, aged between 17 and 87). Each record includes 15 simultaneously measured signals: the conventional 12 leads (i, ii, iii, avr, avl, avf, v1, v2, v3, v4, v5, v6) together with the 3 Frank lead ECGs (vx, vy, vz). Each signal is digitized at 1 kHz with 16-bit resolution over a range of $\pm$ 16.384 mV. To synchronize the sampling frequency with that of the ECG-ID database, all 15 ECG signals from the PTB ECG database was resampled by 500 Hz.

\subsubsection{Face} 
The Faces95 dataset contains 1,440 images (20 images per individual), obtained from 72 individuals (male and female subjects). The individuals are mainly undergraduate students. The image resolution is 180 x 200 pixels (portrait format). During the collection of data, the subjects took a step forward towards the camera. This caused Head scale variation, lighting variation, and translating position of the face in the image. Lastly, there are some expression variations. However, there is no hairstyle variation.

\subsubsection{Fingerprint} 
The FVC2006 contains 7,200 images obtained from 150 individuals. These 7,200 images consist of 4 different sensors, which are Electric Field sensor, Optical Sensor, Thermal sweeping Sensor, and SFinGe v3.0 (1,800 images per sensor). Each database is divided into two subsets, A and B. For subsets DB1-A to DB4-A, each subset contains 140 subjects with 12 images, yielding 1,680 images. For other subsets DB1-B to DB4-B, each subset contains 10 subjects with 12 images, yielding 120 images. The image resolution of the four sensors is 96 x 96 pixels, 400 x 560 pixels, 400 x 500 pixels, and 288 x 384 pixels.

\subsubsection{Virtual Dataset} 
We generated 58 virtual subjects to reduce the variability of the given multimodal database. Each virtual subject is labeled with a person ID and its sex based on ECG and face data. As the subjects of Face95 are mainly undergraduate students, we assumed that they were between their teens and thirties. The gender label is achieved by two annotators fully in agreement. Then, using two criteria: the gender that was labeled on the face data, and the age between teens to thirties, we select a proper sample that considers age and sex from the ECG dataset and match it with face data. Also, we used a fingerprint image from DB1\_A of the FVC2006 database, which are collected with one scanner.

We assigned it randomly to a virtual subject that was already assigned ECG and face data. For face and fingerprint, we augment the data with rotation, translation and cropping method. For ECG, we combine three QRS complexes from 15 segments. Therefore, there are at least 400 data in each biometric for one virtual subject. The model randomly chooses 80\% of the database as training sets, and tests with the remaining data.

\subsection{Experimental Result}

\begin{table}
    \centering
\caption{Accuracy for the Combination of Modality Usage}
\vspace{-0.3cm}
\begin{tabularx}{\columnwidth}{@{}YYYYY@{}}
\hline  \hline
\multicolumn{3}{c}{\textbf{Modality}} & \multicolumn{2}{c}{\textbf{Accuracy (\%)}} \\
\hline
\textbf{ECG} & \textbf{Face} & \textbf{Finger}  & \textbf{ID} & \textbf{Gender} \\
\hline
O &  &   & 77.49 & 91.95  \\
\hline
 & O  &  & 76.44 & 88.51  \\
\hline
 &  & O  & 83.91 & 90.80  \\
\hline
O & O &  & 95.98 & 93.68  \\
\hline
O & & O & 94.83 & 95.40  \\
\hline
 & O & O   & 96.55 & 94.83  \\
\hline
O & O & O & \textbf{98.28} & \textbf{97.70}  \\
\hline  \hline
\end{tabularx}
\label{tab:modality}
\vspace{-0.3cm}
\end{table}

\subsubsection{Unimodal and Multimodal Model}
In this experiment, we add noise for better generalization and discrimination between models. For ECG, we add Gaussian noise for serials of three normalized QRS complexes whose standard deviation is 0.1. For fingerprint, we select 5\% of finger image pixels and change the color to black. For face, we select 97\% of the face image pixels and fill it with black. The changed data are shown in Fig. \ref{fig:noisy}.
In Table \ref{tab:modality}, the accuracy is shown to be more precise for user identification when all three modalities are used (98.28\%) than in all other cases. Additionally, for gender classification, models with all three modalities show better results (97.70\%) than models depending on a sole modality.

From this result, we find that we can determine subjects well with the feature-level fusion model even if input biometrics are noisy. Comparing multimodal models of two modalities with those of three modalities, we see that decisions from multiple biometrics are much more accurate than those from limited biometrics.

\begin{table}
    \centering
\caption{Single Task Model and Multitask Model}
\vspace{-0.3cm}
\begin{tabularx}{0.7\columnwidth}{@{}YYYY@{}}
\hline  \hline
\multicolumn{2}{c}{\textbf{Task}} &  \multicolumn{2}{c}{\textbf{Accuracy (\%)}} \\
\hline
\textbf{ID} & \textbf{Gender} & \textbf{ID} & \textbf{Gender} \\
\hline
O & & 98.28 & -  \\
\hline
 & O & - & \textbf{97.70} \\
\hline
O & O & \textbf{98.97} & 96.55   \\
\hline  \hline
\end{tabularx}
\label{tab:multitask}
\vspace{-0.3cm}
\end{table}

\subsubsection{Single Task and Multitask}
For the experimental result of multitask learning shown in Table \ref{tab:multitask}, we see better results (98.97\% accuracy) in the multitask model for the identification task. It surpasses the performance of a single task model (98.28\%). However, for gender classification, it shows better results when using the single task model (97.70\%). In this situation, we considered the model performance as the averages of both tasks. So, even if we cannot get the best accuracy for gender classification, the performance of the identification task improved. Considering the efficiency in both time and space, the proposed multitask model is much more efficient because its predictions are as good as the single task model while the time required for model training is reduced by half.

\subsubsection{Robustness to Noisy Data}
As shown in Table \ref{tab:noisy}, after adding noise to all biometrics, the highest performances for the two tasks, identification and gender classification, are 98.97\% and 96.55\%, respectively. These experiments illustrate that the performance of multimodal is better than the accuracy of single modality even though noise is included. The result of using clean modalities also indicates similar aspects when using noisy modality. In identification, multimodal has an accuracy of near 100\%. In gender classification, three multimodal provided an accuracy of 99.43\%.
From this experiment, we see that the difference in performance of three multimodal with noise was 1.03\%. This illustrates that the proposed architecture is significantly reliable and robust to noises.

\begin{table}
    \centering
\caption{Multimodal and Multitask Test for Noisy Input}
\vspace{-0.3cm}
\begin{tabularx}{\columnwidth}{@{}YYYYYY@{}}
\hline  \hline
\multicolumn{3}{c}{\textbf{Modality}} &  \multirow{2}{*}{\textbf{Noise}} & \multicolumn{2}{c}{\textbf{Accuracy (\%)}} \\
\cline{1-3} \cline{5-6}
\textbf{ECG} & \textbf{Face} & \textbf{Finger} &  & \textbf{ID}  & \textbf{Gender} \\
\hline
 O & O &   & O & 94.83 & 95.02  \\
\hline
 O &   & O & O & 93.68 & 95.21  \\
\hline
   & O & O & O & 95.21 & 92.91  \\
\hline
 O & O & O & O & 98.97 & 96.55  \\
\hline 
 O & O &   &   & 100.0 & 100.0  \\
\hline 
 O &   & O &   & 98.85 & 96.55  \\
\hline 
   & O & O &   & 100.0 & 98.85  \\
\hline 
 O & O & O &   & 100.0 & 99.43  \\
\hline  \hline
\end{tabularx}
\label{tab:noisy}
\vspace{-0.4cm}
\end{table}

\section{Conclusion}

This paper presents a novel approach for multimodal multitask learning which is robust to noise. ECG, face image, and fingerprint datasets are used for multimodal learning. Additionally, this research focused on user identification and gender classification. The proposed model achieves higher accuracies for both tasks. In addition, the proposed model shows robustness on the spoof attack problem that confronted most models based on single modality. With the results of the experiments, we insist that the performance of the proposed model for user identification and gender classification are better than in previous approaches.

In the future, some supplementary points should be considered to make the proposed model achieve a similar performance when missing one of three biometrics. In addition to user authentication, we are working on improving our model to work with missing modalities. These results would help indicate that it can be deployed in real-world applications with high security. Additionally, the whole network can be trained using an end-to-end approach with more biometrics. Lastly, the technique used for the fusion of multimodal data can be further improved by employing the attention model to choose proper modality or features that are expected to play an important role in the given sample.

\begin{acks}
The authors acknowledge support from the Center for Cyber-Physical Systems, Khalifa University, under Grant Number 8474000137-RC1-C2PS-T3.
\end{acks}

\end{document}